\documentclass[conference]{IEEEtran}
\IEEEoverridecommandlockouts
\usepackage{cite}
\usepackage{amsmath,amssymb,amsfonts}
\usepackage{graphicx}
\usepackage{textcomp}
\usepackage{xcolor}
\usepackage{algorithm}
\usepackage{algpseudocode}
\usepackage{multirow}
\def\BibTeX{{\rm B\kern-.05em{\sc i\kern-.025em b}\kern-.08em
    T\kern-.1667em\lower.7ex\hbox{E}\kern-.125emX}}
\begin{document}

\title{A Comprehensive Review on Noise Control of Diffusion Model\\

% {\footnotesize \textsuperscript{*}Note: Sub-titles are not captured in Xplore and should not be used}

}

\author{\IEEEauthorblockN{Zhehao Guo}
\IEEEauthorblockA{\textit{Information Science} \\
\textit{University of Pittsburgh}\\
Pittsburgh, USA \\
zhg26@pitt.edu}
\and
\IEEEauthorblockN{Jiedong Lang}
\IEEEauthorblockA{\textit{Data Science} \\
\textit{Northeastern University}\\
Boston, USA \\
lang.ji@northeastern.edu}
\and
\IEEEauthorblockN{Shuyu Huang}
\IEEEauthorblockA{\textit{Data Science} \\
\textit{Columbia University}\\
New York, USA \\
sh3967@columbia.edu}

\and
\IEEEauthorblockN{Yunfei Gao}
\IEEEauthorblockA{\textit{Electrical Engineering} \\
\textit{Northwestern University}\\
Evanston, USA \\
yunfeigao2018@u.northwestern.edu}

\and
\IEEEauthorblockN{Xintong Ding}
\IEEEauthorblockA{\textit{Computer Science} \\
\textit{University of California, Berkeley}\\
Berkeley, USA \\
dxt@berkeley.edu}
}

\maketitle

\begin{abstract}

Diffusion models have recently emerged as powerful generative frameworks for producing high-quality images. A pivotal component of these models is the noise schedule, which governs the rate of noise injection during the diffusion process. Since the noise schedule substantially influences sampling quality and training quality, understanding its design and implications is crucial. In this discussion, various noise schedules are examined, and their distinguishing features and performance characteristics are highlighted.

\end{abstract}

\section{Introduction}

Machine learning has experienced exponential growth in recent years, with generative artificial intelligence (AI) emerging as one of its most significant branches. The increasing prevalence of generative AI has led to its widespread integration into various aspects of daily life. A primary application of generative AI is text generation, as demonstrated by platforms such as ChatGPT, which boasts over 180 million users engaging with AI for question answering, conversational interactions, and numerous other tasks. Beyond text generation, the demand for AI-driven image synthesis has been rapidly increasing. Advances in generative AI now enable the transformation of textual descriptions into entirely novel images. Given a set of input sentences, machine learning models can generate unique images that did not previously exist, facilitating creative and design processes across multiple industries. The broad applicability of AI-powered image generation has garnered significant academic and industrial interest, further driving research and development in this field. In the context of image synthesis, two primary types of models—discriminative and generative—can be utilized for static image generation, each with distinct methodologies and capabilities. 
A discriminative model\cite{younis2024machine, sohl2015deep, liu2024visual} is a type of machine learning model designed to differentiate between distinct categories within a given dataset. In contrast, a generative model seeks to learn the underlying data distribution and leverage this knowledge to generate new data samples that resemble the original dataset. In the domain of image generation\cite{croitoru2023diffusion}, generative models have become the predominant statistical approach for research and development\cite{yang2023diffusion}. Various methodologies have been proposed within this field, including Generative Adversarial Networks (GANs), Variational Autoencoders (VAEs), and Flow-based models. These approaches have demonstrated that generative artificial intelligence is capable of producing high-quality images that are perceptually coherent to humans. However, each method presents inherent limitations. Generative Adversarial Networks (GANs) consist of two neural networks: a discriminator and a generator, which are trained through an adversarial process formulated as a minimax game. While GANs have been highly successful in generating realistic images, they are known for challenges such as training instability and limited output diversity. These issues arise due to the adversarial training framework, where the generator and discriminator iteratively optimize against each other. A particularly notable challenge is mode collapse, in which the generator produces a narrow set of outputs rather than fully capturing the diversity of the data distribution. Variational Autoencoders (VAEs) are powerful generative models that integrate probabilistic reasoning with neural networks to learn and generate data from a compressed latent space. This latent space represents a lower-dimensional probabilistic distribution from which the original data can be reconstructed. Unlike other generative models, VAEs utilize a surrogate loss function derived from the Evidence Lower Bound (ELBO) to approximate the data likelihood. While this approach ensures greater training stability, it often comes at the cost of output fidelity, leading to generative results that tend to appear blurrier compared to those produced by Generative Adversarial Networks (GANs). Flow-based models, which facilitate tractable sampling and latent variable inference, employ a series of transformations applied to an initial prior distribution. These models rely on specialized architectural designs that enforce invertibility and reversibility, ensuring the exact computation of the likelihood function through bijective mappings. While these constraints enable precise likelihood estimation, they impose significant limitations on the model’s flexibility and scalability, particularly when dealing with high-dimensional data spaces.

Given the limitations of existing generative models, diffusion models have emerged as the most prominent approach in AI-driven image generation, owing to their greater flexibility and fewer constraints. The introduction of diffusion models has significantly enhanced image generation and editing capabilities, enabling high-quality image synthesis and modification. As their name suggests, diffusion models operate by progressively introducing noise into images during training. Diffusion models can also be employed to generate synthetic data that preserves privacy while maintaining the utility of the original data which benefits for privacy-preserving blockchain applications\cite{song2024unveiling, song2024advancing}. The training process involves gradually corrupting images with noise and subsequently employing a neural network to learn the reverse process, effectively denoising the images to reconstruct the original data distribution. Once trained, the neural network is capable of generating entirely new images from scratch, serving as representations of the learned data distribution. The iterative process of noise addition and removal resembles the physical phenomenon of diffusion, hence the model's name. Unlike other generative approaches, diffusion models belong to a class of probabilistic generative models that transform random noise into structured data samples. A key factor influencing their performance is the noise schedule, which governs the rate at which noise is injected during the training and sampling processes. The choice of an appropriate noise schedule plays a crucial role in determining the quality of the generated images \cite{hoogeboom2023simple, zhang2023adding, parmar2023zero, chang2023muse}, making it a critical parameter in diffusion-based image synthesis.

We begin by detailing the underlying
mechanisms of the diffusion model.

\begin{figure}
    \centering
    \includegraphics[scale=0.15]{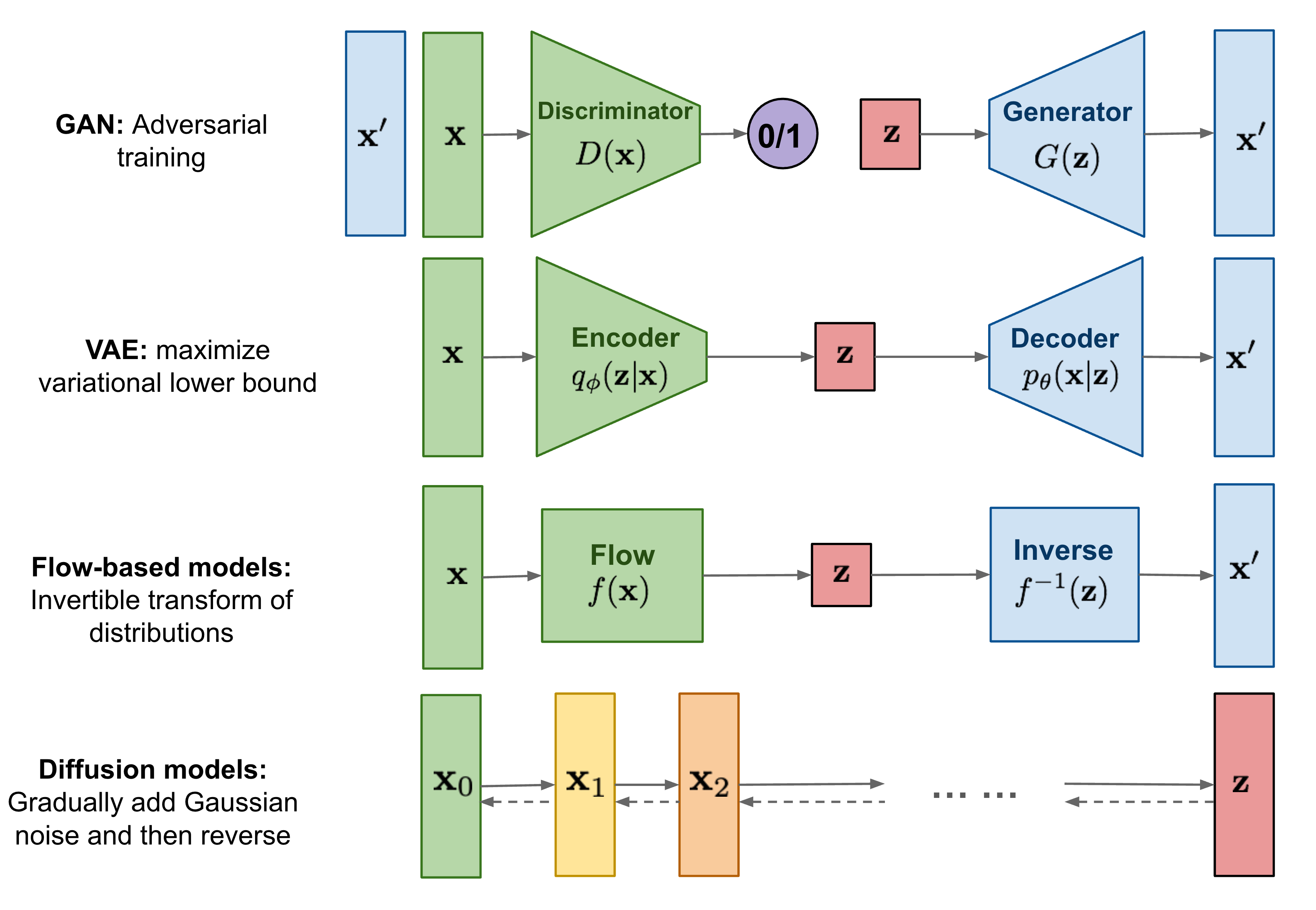}
    \caption{}
\end{figure}

\section{Diffusion Model}
Diffusion models utilize parameterized Markov Chains in their computational framework \cite{saharia2022photorealistic, zou2024segment, khachatryan2023text2video, tang2023emergent}. A Markov Chain represents a sequence of continuous states, where transitions between states occur according to a defined probability distribution. A key characteristic of a Markov Chain is that the next state depends solely on the present state, rather than the entire sequence of preceding states.

In diffusion models, the Markov Chain is leveraged in conjunction with variational inference to facilitate data generation. Variational inference is a technique used to approximate complex probability distributions that are computationally intractable to evaluate directly. During the diffusion process, a neural network is trained to learn the transition dynamics of the reverse diffusion process. The model is trained using Gaussian noise, which is progressively added during the forward process and subsequently removed during the reverse process. This iterative noise injection and removal allow the neural network to reconstruct high-quality data from an initially noisy input, making diffusion models a powerful framework for generative modeling.
\subsection{MATHEMATICS}
The original data sample is written as $x_0$. A sequence of latent variables can be expressed as  $x_0$, $x_1$, $x_2$..., $x_T$, all having the same dimensionality as $x_0$. The whole process is constructed such that $x_T$	follows a Gaussian distribution. The diffusion model denoises $x_T$, which is filled with noise, to reconstruct the original data sample $x_0$. We define q($x_0$) as the real data distribution that forms the data sample $x_0$. 

\subsubsection{FORWARD PROCESS} 
During the forward process, Gaussian noise is incrementally added at each iteration. This process can be effectively represented using a Markov Chain, where each state transition is governed by a predefined probability distribution.
\begin{equation}
\begin{split}
q(x_t \mid x_{t-1}) := \mathcal{N}(x_t; \sqrt{1-\beta_t}\, x_{t-1}, \beta_t I)
\end{split}
\label{eq:eq1}
\end{equation}
Here $\beta$ is a variance schedule that exists in each iteration from $\beta_0$ to $\beta_t$ as $\beta_t \in (0, 1)$. $\beta_t$ is the variance that decides how much noise is added to the $q(x_{t-1})$ to obtain the next data distribution $q(x_t)$. The parameter $\{\beta_t\}$ is called noise schedule, as it controls the rate of noise addition from the original data distribution $q(x_0)$ to an approximately Gaussian distribution $q(x_{t})$\cite{ho2020denoising, chefer2023attend}. $\{\beta_t\}$ increases as t increases. The manner in which $\{\beta_t\}$ increases can take various forms, e.g., linear schedules and exponential schedules. The noise schedule is a critical factor in diffusion models, as it affects both performance and precision\cite{chen2023importance}.
The nosing step from $x_0$ to $x_t$ is the product of all previous Markov chain transitions. The equation can be expressed as below
\begin{equation}
\begin{split}
q(x_1, \ldots, x_T \mid x_0) = \prod_{t=1}^{T} q(x_t \mid x_{t-1})
\end{split}
\label{eq:eq2}
\end{equation}
Moreover, because every step in the forward process leverages Gaussian noise to transition from the previous state of the data distribution to the next step of the data distribution, based on Eq. 2 we can derive an equation from any $q(x_{t})$ to $q(x_0)$:
\begin{equation}
\begin{split}
q(x_t \mid x_{t-1}) = \mathcal{N}\bigl(x_t; \sqrt{1-\beta_t}\, x_{t-1}, \beta_t I\bigr)
\end{split}
\label{eq:eq3}
\end{equation}
\subsubsection{REVERSE PROCESS}
In contrast to the forward process, which progressively adds noise to the data distribution, the reverse process aims to remove noise and reconstruct the original data. This denoising, or reverse, process is facilitated by a deep learning model, which is trained to approximate the noise and ultimately learn to remove it from a Gaussian-distributed latent variable.

 A deep learning learn, parameterized by $\theta$, is designed to learn the reverse noise transformation. By leveraging the transition probability $q(x_t \mid x_{t-1})$, the model can effectively reverse the noise injection process, restoring $x_0$ from $x_t$, where $x_t$ follows an approximate Gaussian distribution. To train the model, an appropriate loss function is minimized, allowing the network to learn to reconstruct the original data distribution or estimate the noise introduced by the forward process\cite{ho2020denoising, wu2023visual}. 
\begin{equation}
\begin{split}
p_\theta(x_{0:T}) = p(x_T) \prod_{t=1}^{T} p_\theta(x_{t-1} \mid x_t)
\end{split}
\label{eq:eq4}
\end{equation}
\subsubsection{LOSS FUNCTION} %ELBO
In the diffusion models, the models are often trained by optimizing a variational bound with the negative log-likelihood. The posterior $q(x_{1:T} \mid x_0)$ is closed to $p_\theta(x_{1:T} \mid x_0)$. The diffusion model leverages Kullback-Leibler(KL) divergence, which is used to measure the similarity between two data distributions, to compare $q(x_{1:T} \mid x_0)$ and $p_\theta(x_{1:T} \mid x_0)$ \cite{ho2020denoising}.
\[
-\log p_\theta(x_0) \;\leq\; 
\]

\[
-\log p_\theta(x_0) 
\;+\; D_{KL}\bigl(q(x_{1:T}\mid x_0)\,\|\,p_\theta(x_{1:T}\mid x_0)\bigr)
\]

\[
=\; \mathbb{E}_{q(x_{1:T}\mid x_0)}\!\biggl[\log\frac{q(x_{1:T}\mid x_0)}{p_\theta(x_{0:T})}\biggr].
\]
Since log$p_\theta(x_0)$ is hard to calculate, the variational lower bound (as known as Evidence Lower Bound, or ELBO) is used to approximate its value. The ELBO consists of a sum of KL divergences between forward and reverse conditional distributions. It is important to note that KL divergence is always larger than or equal to zero, which explains the origin of the term "lower bound". Therefore, by minimizing the loss, we are maximizing the lower bound of the probability of generating real data samples.\cite{sohl2015deep}
When training the Gaussian diffusion models, the loss function is further simplified to enhance training efficiency by removing the weighting component of the original loss function\cite{ho2020denoising, ho2022classifier}. 
\[
L_{\text{simple}}(\theta) := \mathbb{E}_{t,x_0,\epsilon} \biggl[\,\bigl\|\epsilon - \epsilon_{\theta}\bigl(\sqrt{\bar{\alpha}_t}x_0 
+ \sqrt{1 - \bar{\alpha}_t}\,\epsilon,\,t\bigr)\bigr\|^2\,\biggr].
\]
Training with the simplified loss function allows the denoising process to prioritize more challenging tasks, such as removing large amounts of noise at higher values of $t$, while placing less emphasis on simpler tasks, such as eliminating smaller noise levels. This optimization strategy enhances the model's ability to generate higher-quality samples when synthesizing new data \cite{ho2020denoising}.

\section{NOISE SCHEDULE}

As previously discussed, diffusion models are a class of generative artificial intelligence models trained through a process of iterative denoising of random noise. The forward process involves progressively adding noise to the data distribution until the training data converges to a purely Gaussian noise distribution. Conversely, the reverse process seeks to remove this noise to reconstruct the original data.

During both the forward and reverse processes, noise is added or removed in small increments or decrements over a sequence of time steps. The rate of noise addition or removal over time is referred to as the noise schedule, which plays a critical role in the training of diffusion models \cite{sabour2024align}.

An inadequately chosen noise schedule can significantly impact model performance. A slow noise schedule can lead to inefficiencies in training, increasing computational time. Additionally, an improperly selected noise schedule can degrade the quality of generated samples. For instance, applying the same noise schedule to images of different resolutions can lead to poor image reconstruction and information loss\cite{sabour2024align}.

Thus, selecting an appropriate noise schedule is essential for optimizing the performance and efficiency of diffusion model training.  Diffusion models define the noise injection process as following:

\begin{equation}
\begin{split}
x_t = \sqrt{1-\beta_t}\,x_{t-1} \;+\; \sqrt{1 - \beta_t}\,\epsilon_{t-1}
\end{split}
\end{equation}
where $x_0$ represent the input data distribution, and let $\epsilon$ be a sample drawn from a Gaussian distribution. The variable $t$  denotes the timestep in the noise schedule, where $t \in [1, T]$, and $T$ represents the total number of noise schedule steps. The variable $x_t$ corresponds to the data distribution at timestep $t$, while $\beta_t$ is a continuous value within the range $(0, 1)$ expressed as $\{\beta_t \in (0,1)\}_{t=1}^{T}$ \cite{ho2020denoising}. As the t increases, $x_t$ progressively converges toward a pure Gaussian noise distribution.
\subsection{COMMON NOISE SCHEDULE}
Next, we introduce several commonly used noise schedules that are widely employed in the process of diffusion models.
\subsubsection{LINEAR SCHEDULE}
A linear schedule is a noise scheduling strategy in which noise is added or removed at a constant rate throughout the diffusion process. The linear noise schedule strategy is a straightforward approach commonly employed in various diffusion processes, such as DDPM \cite{ho2020denoising}. The variance parameter $\beta_t$ increases uniformly from an initial small value $\beta_0$ at the first timestep to a final value $\beta_T$ at the last timestep. During the reverse process, the noise level decreased linearly. For any $t \in [1, T]$, the noise follows this predefined linear progression.
\begin{equation}
\begin{split}
\beta_t = \beta_1 + \frac{t - 1}{T - 1}(\beta_T - \beta_1)
\end{split}
\end{equation}
where $T$ represents the final timestep, $\beta_T$ denotes the final noise level, $t$ is the current timestep, $\beta_t$ is the current noise level, and $\beta_1$ corresponds to the initial noise level. The complexity of learning the noise removal process gradually increases over time. Initially, when $beta_t$ is small, only a minor amount of noise is introduced into the data distribution. As $beta_t$ increases linearly, the denoising task becomes progressively more challenging, requiring the model to learn how to remove larger amounts of noise at later timesteps.
Although the linear noise schedule is straightforward to implement, it lacks flexibility, as noise is added in a uniform and unadaptive manner, without consideration of the underlying data distribution.

\subsubsection{FIBONACCI SCHEDULE}

A Fibonacci noise schedule is derived from the first $T$ terms of the Fibonacci sequence. The function is defined as follows\cite{chen2020wavegrad}: 

\begin{equation}
\begin{split}
\beta_n = \beta_{n-1} + \beta_{n-2}
\quad \forall\,n \ge 2
\end{split}
\end{equation}

where we define the two numbers of the Fibonacci sequence are defined as follows: $\beta_0 = 1 \times 10^{-6}, 
\quad
\beta_1 = 2 \times 10^{-6}$. In the figure 2\cite{chen2020wavegrad}, in comparison to the linear noise schedule, the Fibonacci noise schedule exhibits a distinct pattern in noise level progression. Initially, the noise level changes gradually; however, at certain points, it decreases significantly, reflecting the characteristic growth pattern of the Fibonacci sequence. Unlike the linear noise schedule, which follows a constant rate of change, the Fibonacci noise schedule is nonlinear and demonstrates a self-similar growth pattern in noise distribution.
\begin{figure}
    \centering
    \includegraphics[scale=0.5]{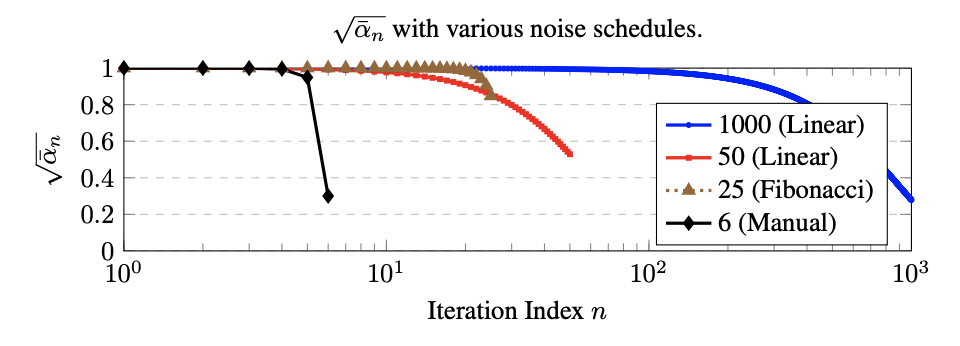}
    \caption{The noise control as time step increases under different types of noise schedule.}
\end{figure}

\subsubsection{COSINE SCHEDULE}
Instead of increasing noise in a linear fashion, a cosine noise schedule employs a smooth, non-linear function that follows a trajectory similar to the cosine function. This scheduling approach is inspired by the mathematical properties of the cosine function, providing a more gradual and adaptive noise progression. With $
\beta_t = 1 - \frac{\bar{\alpha}_t}{\bar{\alpha}_{t-1}}
$, we define COSINE noise schedule in terms of $\bar{\alpha}$ as following function:
\begin{equation}
\begin{split}
\bar{\alpha}_t = \frac{f(t)}{f(0)}
\quad,\quad
f(t) = \cos\!\Bigl(\frac{t/T + s}{1 + s} \cdot \frac{\pi}{2}\Bigr)^2.
\end{split}
\end{equation}
where $s$ is a small constant offset introduced to prevent $\beta_t$ approaching zero at the beginning of the timestep \cite{nichol2021improved}. The relationship between $\beta_t$ and $\bar{\alpha}$ is defined as follows: $\beta_t = 1 - \frac{\bar{\alpha}_t}{\bar{\alpha}_{t-1}}$ \cite{nichol2021improved}. This design ensures that the rate of change in the noise schedule is highest when $\bar{\alpha}$ is at its midpoint, resembling a linear drop-off. However, as the timestep approaches 0 or $T$, the rate of change smooths out, meaning that only a minimal amount of noise is added to the data distribution in these regions. The key distinction between the cosine noise schedule and the linear noise schedule lies in their handling of denoising complexity. The cosine schedule delays the more challenging denoising tasks until after the midpoint of training, which leads to improved sample quality, enhanced training efficiency, and faster convergence\cite{nichol2021improved}. 
\subsubsection{SIGMOID SCHEDULE}
Similar to cosine schedule, the sigmoid noise schedule exhibits a smooth rate of change compared to the linear noise schedule. The sigmoid function transitions gradually from 0 to 1, with its steepest slope occurring around $x=0$. At each timestep $t$, the sigmoid noise schedule is defined by the following function\cite{jabri2022scalable}:
\begin{equation}
\begin{split}
\bar{\alpha}_t \;=\; 
\frac{
-\,\mathrm{sigmoid}\!\Bigl(\tfrac{t\,(\mathrm{e} - \mathrm{s}) + \mathrm{s}}{\tau}\Bigr)
\;+\;
\mathrm{sigmoid}\!\Bigl(\tfrac{\mathrm{e}}{\tau}\Bigr)
}{
\mathrm{sigmoid}\!\Bigl(\tfrac{\mathrm{e}}{\tau}\Bigr)
\;-\;
\mathrm{sigmoid}\!\Bigl(\tfrac{\mathrm{s }}{\tau}\Bigr)
}\,.
\end{split}
\end{equation}
where $s$ represents the starting point of the sigmoid function's range, while $t$ denotes its endpoint. The parameter $\tau$ serves as a temperature coefficient, typically varying around $1$. Selecting an appropriate temperature parameter allows for reducing the noise level's influence towards the end of the diffusion process, thereby optimizing the denoising performance \cite{jabri2022scalable}.
The sigmoid noise schedule begins with a gradual increase and transitions smoothly toward the end. This characteristic enhances the stability of the sigmoid noise schedule and helps mitigate abrupt transitions that could negatively impact sample quality at the beginning and end of training.
For high-resolution images, which exhibit greater redundancy among neighboring pixels, the sigmoid noise schedule demonstrates greater stability compared to the cosine noise schedule \cite{jabri2022scalable}.

\subsubsection{EXPONENTIAL SCHEDULE}

As the name suggests, the exponential noise schedule defines the rate of noise change in a manner that follows an exponential growth pattern. This schedule is designed to ensure that
$\beta_t$ increases exponentially as the timestep $t$ progresses. The function governing the exponential noise schedule is given by:
\begin{equation}
\begin{split}
\beta_t 
\;=\; 
\beta_{\min} 
\Bigl(\frac{\beta_{\max}}{\beta_{\min}}\Bigr)^{\tfrac{t-1}{T-1}}
\quad
\text{for}
\quad
t \in [1,T].
\end{split}
\end{equation}
where $\beta_{\min} = \beta_1$ at $t = 1$, $\beta_{\max} = \beta_T$ at $t = T$. With the function of exponential noise schedule,, the rate of noise increase accelerates rapidly at the beginning compared to the linear and cosine noise schedules. This characteristic makes the exponential noise schedule particularly advantageous in scenarios where the model benefits from the rapid introduction of noise, such as when early-stage features are crucial for effective learning.

\subsubsection{CAUCHY DISTRIBUTION}
The Cauchy distribution is a heavy-tailed probability distribution \cite{hang2024improved} that can be utilized as a noise schedule in the diffusion process. Within the Cauchy distribution framework, the noise schedule is defined as follows:

\begin{equation}
\begin{split}
\beta_t 
  = \text{constant} \times f_{\mathrm{Cauchy}}(t)
\end{split}
\end{equation}
where $f_{\mathrm{Cauchy}}(t) = f(t; x_0, \gamma)
  = \frac{1}{\pi\,\gamma}
    \cdot
    \frac{\gamma^2}{(t - x_0)^2 + \gamma^2}$
 and $\gamma$ represent the scale parameters, which define the width of the peak of the Cauchy distribution. This implies that the noise level reaches its maximum when $t$ is close to $x_0$, as a key characteristic of the Cauchy distribution is that its peak is concentrated near$x_0$, with the noise level decreasing in a heavy-tailed manner as $t$ increases. The Cauchy distribution exhibits a gradual rate of change at the beginning, followed by a sharp transition near the peak. Unlike most conventional noise schedules, such as the cosine schedule, where the noise level changes smoothly and continuously over time without abrupt variations \cite{hang2024improved}, the Cauchy distribution introduces more localized variations.
Additionally, as a heavy-tailed probability distribution, the Cauchy distribution assigns greater weight near its peak while maintaining heavier tails. This property makes the Cauchy distribution a preferred choice when the noise needs to be concentrated within a specific range of time steps, allowing for controlled noise generation over the diffusion process.

\subsubsection{LAPLACE DISTRIBUTION}
The Laplace distribution is a continuous probability distribution that can be utilized as a noise schedule in the diffusion process. It is defined as follows:
\begin{equation}
\begin{split}
f(t; \mu, b)
  = \frac{1}{2b}\,\exp\!\Bigl(-\frac{\lvert t - \mu \rvert}{b}\Bigr)
\end{split}
\end{equation}
where $\mu$ and $b$ are the parameters that determine the center of the peak and the width of the distribution, respectively. A Laplace distribution noise schedule concentrates the majority of noise injection around its center, meaning that more noise is introduced during mid-range timesteps. Since a greater proportion of noise is generated at mid-range timesteps, this scheduling approach can be computationally efficient, offering improved performance under resource-constrained conditions \cite{hang2024improved}.

\subsubsection{LOGISTIC SCHEDULE}
 Like the cosine noise schedule, the logistic noise schedule exhibits a smooth rate of change compared to the linear noise schedule. It maintains a gradual transition over discrete timesteps $t$, ranging from 1 to $T$. A key characteristic of the logistic schedule is that it introduces noise slowly at the beginning, followed by a rapid increase in the noise change rate around the midpoint of training \cite{lin2024schedule}. This structure helps preserve critical features of the training data. As $t$ approachs $T$, the rate of noise increase gradually smooths out, mirroring the behavior observed at the start of the process. The logistic noise schedule can be expressed as follows:
\begin{equation}
\begin{split}
\bar{\alpha}_t = \frac{1}{1 + e^{-k\,(t - t_0)}}.
\end{split}
\end{equation}
where $k$ and $t_0$ are hyperparameters that govern the behavior of the logistic noise schedule. The parameter $k$ is a steepness coefficient that controls how the curve evolves over time, determining how rapidly the function transitions from near 0 to near 1. Meanwhile, $t_0$ defines the midpoint of the curve, representing the timestep at which the transition from low to high noise levels occurs, typically around 50\% of the curve's progression.
The logistic noise schedule begins with a gradual increase in noise and transitions smoothly towards the end. This characteristic enhances stability and helps mitigate abrupt transitions that could negatively impact sample quality at both the beginning and end of training.  By preserving the integrity of the original data distribution $x_0$, this approach contributes to improved training stability and overall generative performance.
As shown in figure2\cite{lin2024schedule}, the noise value varies across different types of noise schedules. here, $\bar{\alpha}_t$ applies reparameterization \cite{ho2020denoising} to express $x_t$ at any timestep $t$ in terms of $x_0$ and $\beta$. In this parameterization, $\alpha_t$ is introduced as a replacement for $\beta$ to facilitate the transformation process. The expressions are listed as following: 
$\alpha_t := 1 - \beta_t
\quad \text{and} \quad
\bar{\alpha}_t := \prod_{s=1}^t \alpha_s.$
\begin{equation}
\begin{split}
\mathbf{x}_t = \sqrt{\bar{\alpha}_t}\,\mathbf{x}_0 
              + \sqrt{1 - \bar{\alpha}_t}\,\boldsymbol{\epsilon}
\end{split}
\end{equation}
The figure2 illustrates a linear change in noise levels during the midpoint of the diffusion process. By adhering to the continuous dynamics of the diffusion process, this approach minimizes unintended deviations and reduces errors, resulting in more accurate and stable latent predictions. Consequently, it improves the overall fidelity of the inversion process \cite{lin2024schedule}.

\subsubsection{MONOTONIC NEURAL NETWORK}
Instead of employing a fixed noise schedule, a learned noise schedule can be utilized, specifically one modeled by a monotonic neural network with a sigmoid-like function \cite{kingma2021variational}. The noise schedule is optimized through the following parameterization: 
\begin{equation}
\begin{split}
\sigma_t^2 = \operatorname{sigmoid}\bigl(\gamma_{\eta}(t)\bigr)
\end{split}
\end{equation}
where $\gamma_{\eta}(t)$ represents a monotonic neural network parameterized by $\eta$, which satisfies the boundary condition SNR(t) $<$ SNR(s) for any t $>$ s. The neural network consists of three linear layers with output dimensions of 1, 1024, and 1, respectively. The equation $\alpha_t = \sqrt{1 - \sigma_t^2}$ is employed to simplify the parameterization, leading to the formulations: $\alpha_t^2 = \mathrm{sigmoid}\bigl(-\gamma_{\eta}(t)\bigr)$ and $\mathrm{SNR}(t) = \exp\bigl(-\gamma_{\eta}(t)\bigr)$ \cite{kingma2021variational}. This formulation enables the noise schedule to become a set of learnable parameters, allowing diffusion models to dynamically optimize the manner in which noise is added over time. This approach not only leads to lower variance estimates, resulting in faster and more stable training compared to fixed, handcrafted noise schedules \cite{kingma2021variational, sahoo2023diffusion}, but it also excels at generating robust representations of data and can be used to enhance or denoise feature extraction from sensor inputs. For instance, it leads to improving landmark detection and reliable keypoint matching \cite{10503743, 10684902, qu2024visual}.

\section{CONCLUSION}
The noise schedule plays a critical role in the diffusion process. Thus, selecting an appropriate noise schedule can significantly enhance the performance and quality of diffusion-based generative models \cite{chen2023importance, chen2023generalist}. However, there is no universal noise schedule that is optimal for all diffusion processes. Instead, different noise schedules perform more effectively under specific conditions.
For instance, the sigmoid noise schedule demonstrates superior performance compared to the cosine noise schedule when applied to high-resolution images. Additionally, the impact of noise schedules varies not only across different scenarios but also at different stages of the diffusion process. For example, during training, the sigmoid noise schedule provides greater stability than the cosine noise schedule. However, during the sampling stage, the difference in performance between these two schedules is less significant \cite{jabri2022scalable}.
Moreover, even within the same noise schedule, parameter adjustments can lead to substantial variations in performance. As shown in Table 1 \cite{chen2023importance, karras2022elucidating}, the effectiveness of a noise schedule can vary based on different parameter settings. Therefore, to achieve optimal performance in the diffusion process, it is essential to explore different noise schedules and fine-tune their parameters accordingly. 

\begin{table}[h]
\centering
\caption{Noise schedule function $\gamma(t)$. The performance comparison of same noise schedule with different parameters.}
\begin{tabular}{lccc}
\textbf{Noise schedule function $\gamma(t)$} & 
\textbf{64$\times$64} & 
\textbf{128$\times$128} & 
\textbf{256$\times$256}\\
$1 - t$ & 2.04 & 4.51 & 7.21 \\
\textbf{cosine} $(s=0,e=1,\tau=1)$ & 2.71 & 7.28 & 21.6 \\
cosine $(s=0.2,e=1,\tau=1)$ & 2.15 & 4.90 & 12.3 \\
cosine $(s=0.2,e=1,\tau=2)$ & 2.84 & 5.64 & 5.61 \\
cosine $(s=0.2,e=1,\tau=3)$ & 3.30 & 4.64 & 6.24 \\
\textbf{sigmoid} $(s=-3,e=3,\tau=0.9)$ & 2.09 & 5.83 & 7.19 \\
sigmoid $(s=-3,e=3,\tau=1.1)$ & \textbf{2.03} & 4.89 & 7.23 \\
sigmoid $(s=0,e=3,\tau=0.3)$ & 4.93 & 6.07 & 5.74 \\
sigmoid $(s=0,e=3,\tau=0.5)$ & 3.12 & 5.71 & 4.28 \\
sigmoid $(s=0,e=3,\tau=0.7)$ & 3.34 & \textbf{3.91} & 5.49 \\
sigmoid $(s=0,e=3,\tau=0.9)$ & 2.29 & 4.42 & 5.48 \\
sigmoid $(s=0,e=3,\tau=1.1)$ & 2.36 & 4.39 & 7.15 \\
\end{tabular}
\end{table}

\bibliographystyle{IEEEtran}
\bibliography{references}

\end{document}